%% file: main.tex
\documentclass[sigconf, nonacm]{acmart}

\usepackage{verbatim}
\usepackage{tikz}
\usetikzlibrary{graphs}
\usetikzlibrary[quotes]
\usetikzlibrary{backgrounds}
\usepackage{tikz-qtree}
\usepackage{multirow}
\usepackage{balance}
\usepackage{amsmath}
\usepackage{etex}
\usepackage{tabu}
\usepackage{booktabs}
\usepackage{comment}
\usepackage{subcaption}
\usepackage{color,soul}
\usepackage{paralist, tabularx}
\usepackage{url}
\usepackage{changepage}
\usepackage{makecell}

 \usepackage{algorithmicx,algorithm}
 \usepackage[noend]{algpseudocode}
\usepackage{enumitem}

\newcommand{\pbm}{\textsc{Eda4Sum}}

\newcommand{\rlsum}{\textsc{RLSum}}
\newcommand{\topsum}{\textsc{Top1Sum}}
\newcommand{\rlsumic}{\textsc{RLSum\_IC}}
\newcommand{\rlsumdc}{\textsc{RLSum\_DC}}

\newcommand{\rlsumhu}{\textsc{RLSum\_HU}}

\newcommand{\rlsumln}{\textsc{RLSum\_LN}}

\newcommand{\rlsumbl}{\textsc{RLSum\_BL}}

\newcommand{\topsumhu}{\textsc{Top1Sum\_HU}}

\newcommand{\reva}[1]{{\leavevmode\color{black}{#1}}}
\newcommand{\revb}[1]{{\leavevmode\color{black}{#1}}}
\newcommand{\revc}[1]{{\leavevmode\color{black}{#1}}}
\newcommand{\common}[1]{{\leavevmode\color{black}{#1}}}

\newtheorem{theorem}{Theorem}[section]

\newtheorem{problem}[theorem]{Problem}

\def\Item$$#1$${ $\displaystyle#1$
	\hfill\refstepcounter{equation}(\theequation)}

\def\HiLi{\leavevmode\rlap{\hbox to \hsize{\color{red!20}\leaders\hrule height .8\baselineskip depth .5ex\hfill}}}

\newcommand{\sysName}{\textsc{EDA4Sum}}

\newcommand{\byfacet}{\textsf{by-facet}}

\newcommand{\bysuperset}{\textsf{by-superset}}

\newcommand{\byneighbors}{\textsf{by-neighbors}}
\newcommand{\bydistrib}{\textsf{by-distrib}}

\let\oldnl\nl
\newcommand{\nonl}{\renewcommand{\nl}{\let\nl\oldnl}}

\title{Guided Exploration of Data Summaries}

\author{Brit Youngmann}
\affiliation{%
  \institution{MIT CSAIL}
    \country{USA}
}
\email{brity@mit.edu}

\author{Sihem Amer-Yahia}
\affiliation{%
  \institution{CNRS, Univ. Grenoble Alpes}
    \country{France}
}
\email{sihem.amer-yahia@cnrs.fr}

\author{Aurélien Personnaz}
\affiliation{%
  \institution{CNRS, Univ. Grenoble Alpes}
     \country{France}
}
\email{aurelien.personnaz@cnrs.fr}

\begin{document}
\begin{abstract}
Data summarization is the process of producing interpretable and representative subsets of an input dataset. It is usually performed following a one-shot process with the purpose of finding the best summary. A useful summary contains $k$ {\em individually uniform} sets that are {\em collectively diverse} to be representative. Uniformity addresses interpretability and diversity addresses representativity. Finding such as summary is a difficult task when data is highly diverse and large.
We examine the applicability of Exploratory Data Analysis (EDA) to data summarization and formalize \pbm, the problem of guided exploration of data summaries that seeks to sequentially produce \common{connected summaries} with the goal of maximizing their cumulative utility. \pbm\ generalizes one-shot summarization. We propose to solve it with one of two approaches: (i) \topsum\ that chooses the most useful summary at each step; (ii) \revb{\rlsum\ that} trains a policy with Deep Reinforcement Learning that rewards an agent for finding a diverse and new collection of uniform sets at each step. We compare these approaches with one-shot summarization \common{and top-performing EDA solutions}.
We run extensive experiments on \reva{three} large datasets. Our results demonstrate the superiority of our approaches for summarizing very large data, and the need to provide guidance to domain experts.
\end{abstract}

\maketitle



\input{intro}

\input{related}

\input{model}
\input{End2EndSum}

\input{exp}

\input{conc}


\balance
\bibliographystyle{ACM-Reference-Format}
\bibliography{bibtex}

\end{document}

%% file: intro.tex
\section{Introduction}
\label{sec:intro}
The goal of data summarization is to produce a smaller and informative dataset \cite{wen2018interactive,el2014interpretable} from an input dataset. That is usually achieved by seeking interpretable and representative subsets of the input. Intuitively, a useful summary contains $k$ {\em individually uniform} sets that are {\em collectively diverse} to be representative \cite{swap, wen2018interactive}. Uniformity addresses interpretability as it allows to produce a description for each set, and diversity addresses representativity by seeking to cover data variety.
\reva{This is particularly important for large and highly diverse datasets such as the Sloan Digital Sky Survey (SDSS), a database commonly used in the astrophysics community~\cite{SDSS}. }
\revc{SDSS includes galaxies that belong to $169$ classes defined by the Galaxy Zoo classification \cite{Willett2013}.  
In SDSS, each galaxy has $7$ attributes describing magnitude in each color filter (the attributes $u,g,r,i$, and $z$), size (the attribute
{\em petroRad\_r}), and how far a galaxy is from the Earth (the attribute {\em redshift}).}
A single one-shot summary of SDSS is not representative. 
Indeed, today's astrophysicists spend considerable time running SQL queries
against the SkyServer database. Most of their time is spent in reformulating
queries, searching for galaxy sets
with similar properties or value distributions. In this paper, we investigate the applicability of Exploratory Data Analysis (EDA) to summarizing such large data. 

\begin{figure}
    \centering
    \includegraphics[width=\linewidth]{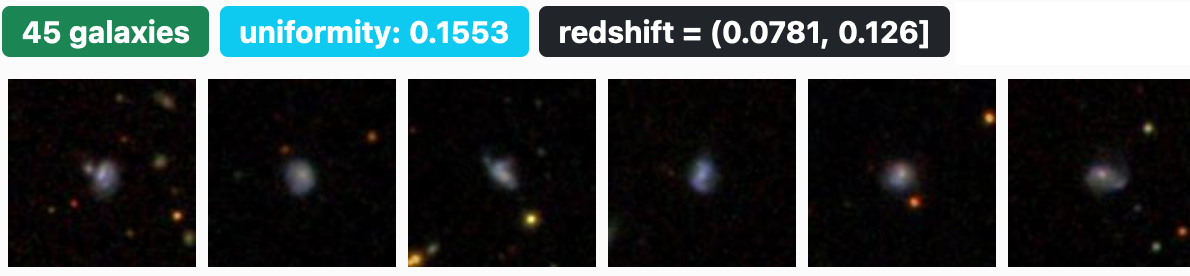}
    \includegraphics[width=\linewidth]{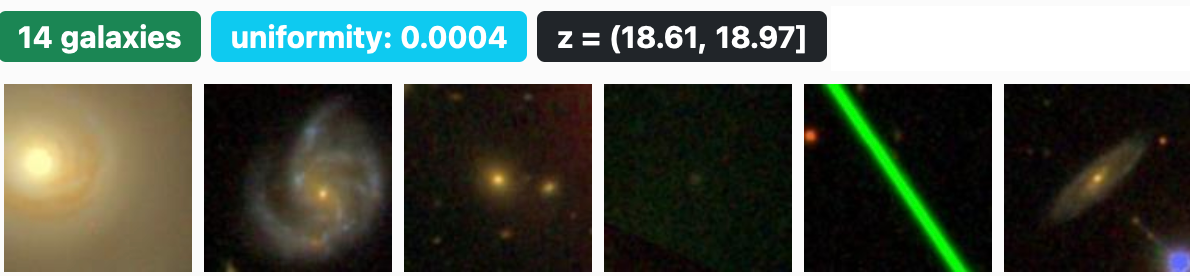}
    \vspace{-8mm}
    \caption{\revc{Examples of uniform (top) and non-uniform (bottom) galaxy itemsets.}}
    \label{fig:example}
    \vspace{-4mm}
\end{figure}

A summary can be defined as a diverse set of $k$ sets of items (referred to
as \emph{itemsets}), each of which is uniform, i.e., it contains items that are
similar to each other. Itemsets are different from each other, leading to a diverse summary. 
Figure \ref{fig:example} shows examples of uniform
and non-uniform itemsets of galaxies derived from SDSS. One can
see that uniform itemsets are easier to interpret by humans. 
Users may not be easy to consume a large amount of information in one step (i.e., shot). Thus, a one-shot summarization approach that leverages a diversity algorithm to find the top-$k$ most uniform and diverse sets appears as a natural solution. 
For instance, SWAP, a common diversity algorithm \cite{swap} would greedily finds $k$
most diverse itemsets subject to a threshold on utility (uniformity
in this case) and is shown to enjoy good approximation guarantees.
\reva{However, with a large and diverse dataset, a one-shot summary may not suffice to cover the variety of itemsets in the database.} For instance, in the case of SDSS, even a summary of 10 itemsets would not cover the 169 galaxy types it contains. 
In fact, with all diversity algorithms
(SWAP \cite{swap}, MMR \cite{mmr}, and GMM \cite{gmm}), there is a tension between
displaying $k$ uniform itemsets and covering data variety. This motivates the
use of EDA for data summarization.

Rather than aiming to cover the entire data in a single one-shot summary \cite{chandola2007summarization,kim2020summarizing}, we seek to tackle a more general problem that adopts a multi-shot approach to summarization where each step produces the most uniform and diverse itemsets, \common{and the collection of steps forms a connected set of summaries.} \reva{To achieve that, we must address two challenges: define summary utility as a function of uniformity and diversity, and make sure the generated summaries at each step are both new and related to previous summaries, to preserve the train of thought of the user~\cite{DBLP:conf/ijcai/ShahafG11}.} This gives rise to the \pbm\ Problem that seeks to find a sequence of summaries whose \emph{cumulated utility} is maximized (Section \ref{sec:model}). \reva{While some datasets are largely uniform, others contain more diversity. Therefore, we need to make sure our optimization function provides tunable weights for uniformity and diversity.}
Generated summaries are operation-driven - unlike previous work where summaries are data-driven \cite{wen2018interactive}. Each step is triggered by an EDA operator that takes an itemset (from the current summary) and returns (at most) $k$ itemsets (i.e., a summary). 
The pipeline has a fixed length, and when it is equal to 1, the problem reduces to a one-shot summarization. We prove that the \pbm\ Problem is NP-hard by a reduction to the Heaviest Path problem in a weighted directed graph \cite{schrijver2003combinatorial}. Thus there is a need for efficient and scalable algorithms to solve it. 


We bootstrap a summarization pipeline by running the SWAP algorithm. 
Thus, for a one-shot summary (i.e., summarization pipeline of length $1$), \sysName\ reduces to SWAP. Other one-shot summarization algorithms could be used to generate the first step.
For subsequent steps, we propose \topsum\ and \rlsum, \common{two adaptations of existing algorithms}. At each step, the algorithm picks one of the itemsets returned by the previous step and chooses which operator to execute on that itemset, resulting in a new summary. 
\topsum\ is a greedy algorithm that chooses to return the highest utility summary at each subsequent step.
Following existing work in EDA~\cite{tovasigmod20,DBLP:journals/pvldb/SeleznovaOAS20,DBLP:conf/sigmod/PersonnazABFS21} we investigate the applicability of Deep Reinforcement Learning to summarization. We design \rlsum, an adaptation of Deep Reinforcement Learning to simulate an agent that learns an end-to-end summarization policy as a sequence of EDA operators that yield the highest reward. 

We ran experiments on \reva{three large and different datasets}. 
\revc{We examined the utility of returned summarization pipelines and found that in most cases (and as expected), \topsum\ returns higher utility summaries than \rlsum. However, \rlsum\ is at least one order of magnitude faster than \topsum.
A specific use case also showed that \rlsum\ performs better than \topsum\ in finding ground-truth itemsets. }
We also examine the use of different EDA operators in building summarization pipelines. 
We find that the traditional drill-down and roll-up operators are not suffice for finding useful summaries, especially in SDSS that requires more expressive operators to cover the variety of galaxy types.
\revc{We also ran an experiment to validate our reward function and found that it outperforms baseline DRL with familiarity and curiosity~\cite{DBLP:conf/sigmod/PersonnazABFS21}.}
Finally, an investigation with two domain expert astronomers who are familiar with SDSS revealed the benefit of using partial guidance for summarization.

%% file: related.tex
\section{Related Work}
\label{sec:related}

We refer to our system as \sysName. Table~\ref{tab:relatedwork} summarizes the differences between \sysName\ and previous work. Columns in bold highlight our novelty, namely: \common{\sysName\ enables generating a \emph{multi-step} summarization pipeline. Each summary in this pipeline is \emph{connected} to the previous summary. This is done by applying exploration \emph{operators which dictate the next summary} to be displayed. \sysName\ also enables \emph{fully automate} generation of summarization pipelines}.
We now describe multiple lines of work that are relevant to ours.

\begin{table*}[t]
\small
\centering
\caption{\revc{Positioning of \sysName\ with respect to Data Exploration and Result Summarization and Explanation.}}
\label{tab:relatedwork}
\vspace{-3mm}
\begin{tabular}{|c|c|c|c|c|c|c|c|}
    \hline
 
\multicolumn{2}{|c|}{\multirow{2}{*}{Related Work}}  &   \multicolumn{2}{c|}{\revc{Pipeline}} &   \multicolumn{2}{c|}{Recommendation} &   \multicolumn{2}{c|}{Guidance} \\\cline{3-8}

\multicolumn{2}{|c|}{}                    &   One-Shot    &\textbf{ Multi-Step}     &    Data-Driven      & \textbf{Operation-Driven}  & \revc{Hands-Free}    &   {\bf \revc{Connected}} \\
    \hline
\multirow{1}{*}{EDA} & \cite{amer2021exploring,amer2021subdex,bar2020automatically,DBLP:journals/pvldb/SeleznovaOAS20}
 &&\textbf{\checkmark}&\textbf{\checkmark}&&\textbf{\checkmark}&\textbf{\checkmark}\\
 &\cite{DBLP:conf/sigmod/PersonnazABFS21, DBLP:conf/cikm/PersonnazABFS21}&&\textbf{\checkmark}&&\textbf{\checkmark}&\textbf{\checkmark}&\textbf{\checkmark}\\
 &\cite{milo2018next,joglekar2017interactive,siddiqui2016effortless,amer2017exploring,wongsuphasawat2015voyager}&&\textbf{\checkmark}&\textbf{\checkmark}&&&\textbf{\checkmark}\\

 &\cite{lee2019avoiding}&\textbf{\checkmark}&&\textbf{\checkmark}&&&\\

 &\cite{dibia2019data2vis}&\textbf{\checkmark}&&\textbf{\checkmark}&&\textbf{\checkmark}&\\

    \hline
 \multirow{1}{*}{\begin{tabular}{@{}c@{}}Summarization \\ and Explanation\end{tabular}} & \cite{wen2018interactive,vollmer2019informative,el2014interpretable,agarwal2007efficient,sarawagi1999explaining}
 &\textbf{\checkmark}&&\textbf{\checkmark}&&&\textbf{\checkmark}\\
 
& \cite{swap}&\textbf{\checkmark}&&\textbf{\checkmark}&&\textbf{\checkmark}&\\

&\cite{kim2020summarizing}&\textbf{\checkmark}&&\textbf{\checkmark}&&\textbf{\checkmark}&\textbf{\checkmark}\\

    \hline
  \multirow{1}{*}{EDA for Summarization} & \sysName
 &\textbf{\checkmark}&\textbf{\checkmark}&\textbf{\checkmark}&\textbf{\checkmark}&\textbf{\checkmark}&\textbf{\checkmark}\\




    \hline
\end{tabular}
\vspace{-4mm}
\end{table*}

\textit{One-shot data summarization}:
A large variety of approaches have been
proposed for summarizing data \cite{kim2020summarizing}. Prominent examples include approaches based
on the Minimum Description Length \cite{bu2005mdl,lakshmanan2002generalized}, approaches that identify extreme aggregates
\cite{wen2018interactive,sathe2001intelligent}, methods that summarize all aggregates \cite{lakshmanan2002quotient}, and works that produce
$k$ diverse clusters showing
common properties in the data \cite{DBLP:conf/sigmod/RoyACDY10}.
Unlike our work, all the methods mentioned above consider data summarization as a one-shot task. 

Approaches that summarize all data typically trade-off summary size against
information loss \cite{kim2020summarizing}.
As mentioned in the Introduction, in cases where the data size is massive, finding the most uniform and diverse parts is helpful. Thus, unlike previous works, since our goal is not to summarize the entire input, we do not measure summary quality in terms of information loss. 
Therefore, we define summarization as the task of finding the most uniform and diverse subsets of the data.
A natural solution to this definition is to leverage diversity algorithms such as SWAP~\cite{swap}, MMR~\cite{mmr}, GMM~\cite{gmm}, and QAGView~\cite{wen2018interactive}. However, as discussed in the Introduction, there is a tension between displaying $k$ sets and covering variety in data. 


\vspace{1mm}
\textit{Result diversification}: 
Result diversification is well-studied in query answering in databases \cite{10.14778/2350229.2350233}, search engines \cite{gollapudi2009axiomatic,guy2021improving} and recommender systems \cite{yu2009takes}.
This problem aims to return $k$ results that take both utility and diversity into consideration
\cite{drosou2017diversity}. In many cases, diversity comes at the cost of utility \cite{ziegler2005improving,10.14778/2350229.2350233}. A common approach to measuring diversity, which we also adopted in our work, relies on pairwise similarities \cite{fraternali2012top, 10.14778/2350229.2350233}.
The main departure from previous work is that we also account for novelty among itemsets selected in previous steps, and the that the score of novelty may change along the summarization pipeline (see Section \ref{sec:impl}).


\vspace{1mm}
\textit{Multi-step data exploration}: 
Data exploration is a multi-step process whose goal is to extract insights from data \cite{milo2018next,marcel2019towards,zhou2020table2analysis}. 
Guiding users in performing data exploration is a well-studied task \cite{friedman1974projection,bar2020automatically,amer2021exploring}.
Numerous works proposed next-step recommendations \cite{milo2018next,eirinaki2013querie,dimitriadou2016aide, huang2018optimization,amer2021subdex}. 
Novel operators for interactively exploring data and discover interesting sets of tuples were introduced in \cite{joglekar2017interactive,lee2019avoiding}.


As opposed to this line of work whose goal is to extract general insights, our goal is to summarize massive datasets by detecting highly uniform and diverse itemsets. Also, as can be seen in Table \ref{tab:relatedwork}, our next-step recommendations are operation-driven - unlike previous work where recommendations are data-driven. This allows generating pipelines that exploit semantic relationships between data regions and preserve the train of thought of the user~\cite{DBLP:conf/ijcai/ShahafG11}.

\vspace{1mm}
\textit{ML for data exploration}: 
Recent work suggested to automate data exploration  using Reinforcement Learning \cite{bar2020automatically,DBLP:journals/pvldb/SeleznovaOAS20,DBLP:conf/sigmod/PersonnazABFS21, DBLP:conf/cikm/PersonnazABFS21}. \sysName\ adopts a similar approach to provide guidance to users with no need for training data. The logic of our \rlsum\ algorithm is based on the system presented in \cite{DBLP:conf/sigmod/PersonnazABFS21, DBLP:conf/cikm/PersonnazABFS21} that guides users
in finding items of interest in large datasets.  
In this system, the process is driven by data familiarity and curiosity.
Unlike \cite{DBLP:conf/cikm/PersonnazABFS21}, \rlsum\ does not require
an extrinsic reward, alleviating the need for labeled data or
prior knowledge.
Moreover, in \rlsum\ the iterative summarization process is driven by uniformity, diversity, and novelty.

%% file: model.tex
\section{Data Model}
\label{sec:model}
We consider a set of items $D$ described with a set of (numerical or categorical) ordinal attributes $A$. Without loss of generality, we will use SDSS to illustrate our data model. 
Numerical attribute values are assumed to be binned into a fixed number of bins. Each item $d {\in} D$ is represented as a vector, denoted as $v_d$, where an entry $v_d(a)$ is the value of $d$ for an attribute $a {\in} A$. 

Following \cite{DBLP:journals/pvldb/SeleznovaOAS20}, we use the notion of \emph{itemset} defined as a set of items that share the same values for a set of attributes. Those attributes define the \emph{itemset description} that has the benefit of conveying the content of the itemset at a glance. 
$\mathcal{D}$ denotes the set of all itemsets created from $D$. \reva{We note that itemsets may overlap}.
To illustrate, Figure~\ref{fig:example} 
contains examples of galaxy itemsets along with their descriptions. We represent each \reva{itemset $i$ with a vector $v_i$ that computes the aggregated values of items in $i$} \emph{for each attribute in $A$}. The value of each vector entry is computed as the mean of the values of its corresponding attribute in the itemset. Other aggregations could be used (e.g., median for ordinal attributes). 



\subsection{Data Summaries}
\label{subsec:summary}
\reva{A summary $I {\subseteq} \mathcal{D}$ is a set of (at most) $k$ itemsets in $\mathcal{D}$}, where $k$ is a system parameter. Intuitively, a useful summary contains itemsets consisting of similar items (uniformity), and where itemsets are pairwise different from each other (diversity). Since our aim will be to generate multi-step summaries, an important question is to what extent the current step's summary displays new itemsets when compared to previous steps' summaries (novelty). 
To that end, we define the notions of uniformity, diversity and novelty of a summary, to be used to define the \emph{utility of a summary}. 

Uniformity of a summary measures how similar items are to each other in each of its itemset. We first define the uniformity of an itemset.
Let $var_a(i)$ denote the variance of items in an itemset $i$ w.r.t. an attribute $a$, where $var$ is some variance measure:
\setlength{\abovedisplayskip}{0pt}
\setlength{\belowdisplayskip}{0pt}
\begin{align*}
uni(i) := \frac{|A|}{\sum_{a \in A} var_a(i)}
\end{align*}


The uniformity of a summary of $k$ itemsets $I$ is given by: 
\setlength{\abovedisplayskip}{0pt}
\setlength{\belowdisplayskip}{0pt}
\begin{align*}
 \mathit{Uni}(I) = \min_{i \in I} (\mathit{uni}(i))
 \end{align*}





The diversity of a summary, denoted as $\mathit{Div}(I)$, is defined as:
$$\mathit{Div}(I) := \min_{i, i' \in I, i <> i'} \mathit{vectorDist}(v_i,v_{i'})$$
where $\mathit{vectorDist}(v_i,v_{i'})$ is the distance between itemsets $i$ and $i'$.



Let $\mathit{SEEN}$ denote the set of all itemsets seen by the user. Initially, $\mathit{SEEN} {=} \emptyset$. \reva{Whenever an itemset is displayed, we add it to the set $\mathit{SEEN}$}. Intuitively, the novelty of a summary captures the proportion of how many new itemsets the user is currently seeing. 
The novelty of a summary $I$ is defined as:
$Nov(I,\mathit{SEEN}) := \frac{I \setminus \mathit{SEEN} }{|I|}$.

\reva{Our objective is to balance uniformity, diversity, and novelty. Following common approaches for results diversification \cite{fraternali2012top, mmr}, we propose a parameterized objective which enables users to specify their desired balance.}
The utility of a summary $I$ is denoted as $utility(I)$, and is defined as follows: 
\setlength{\abovedisplayskip}{0pt}
\setlength{\belowdisplayskip}{0pt}
\begin{align}\label{utility}
      utility(I) {=} \alpha {\cdot} \mathit{Uni}(I) {+} \beta {\cdot} \mathit{Div}(I) {+} \gamma {\cdot} \mathit{Nov}(I,\mathit{SEEN}) 
\end{align}
where $\alpha, \beta , \gamma \in [0,1]$ are system parameters, and $\alpha + \beta + \gamma = 1$.

\reva{We examined the impact of these parameters on the results, determining the ranges in which no change in performance was observed. We found that with a slight change of the parameters values, the cumulated utility
is unaffected. Thus, to allow for a  user-friendly use of \sysName, we reduced the space of all possible combinations of values for the parameters $\alpha, \beta$, and $\gamma$ by allowing each parameter to take either a low, medium, or a high value. We consider the following four combinations of values: (i)-(iii) one of the parameters is set to a high value with the rest assigned to low; (iv) all parameters are set to have a medium value.}

\reva{While some datasets are largely uniform, others contain more diversity. Therefore, we need to make sure our optimization function provides tunable weights for uniformity, diversity and novelty.}


\subsection{Problem Statement}
\label{subsec:problem}
\reva{A single one-shot summary of a large dataset is not representative. We propose to examine the applicability of EDA to data summarization. We first define summarization pipelines whose aim is to generate connected summaries. 
At each step, the user sees a summary $I'$ that is obtained by applying an exploration operator, $explore()$, to an itemset chosen from a summary that was shown in the previous step. The application of an operator to generate the next summary helps understand links between consecutive summaries, and preserve the stream of consciousness of users~\cite{DBLP:conf/ijcai/ShahafG11}.}
A summarization pipeline is a sequence of summaries, connected by exploration operators. In its general form, an operator, denoted as $explore(i,k)$, takes an itemset $i$ and a number $k$, and returns a summary formed by (at most) $k$ itemsets that are related to items in $i$. The pipeline is bootstrapped with various diversity algorithms to start with a summary containing the most uniform and diverse sets.  
In Section \ref{sec:impl}, we describe the exploration operators we support.



Given a bound $t$, the system produces a summarization pipeline of length $t$.
We define the utility of a step to be the utility of its resulting summary. 
The system needs to 
decides which summary to display at each step to maximize the \emph{cumulated utility}. We refer to this question as the \pbm\ Problem and formalize it as follows. 

\begin{problem}[The \pbm\ Problem]
Given an input itemset $i {\in} \mathcal{D}$ and a bound on the number of steps $t$, recommend a summarization pipeline $(I_1,\ldots,I_t)$ of length $t$ with the highest cumulated utility, i.e., $\sum_{j = 1}^t utility(I_j)$ is maximized, and for every $j {\in} [1,t]$, the summary $I_{j+1}$ is obtained by applying an $explore$ operator on an itemset $i$ from the summary $I_j$.  
\end{problem}

We can prove that the \pbm\ Problem is NP-hard by a reduction to the Heaviest Path problem in weighted directed graphs~\cite{schrijver2003combinatorial}.

%% file: End2EndSum.tex
\section{Algorithms}
\label{sec:algos}
\common{To address the \pbm\ Problem, we develop \topsum\ and \rlsum, two adaptations of well-known approaches.} Our algorithms are integrated into a prototype system that is also called \sysName. 

\paragraph*{Architecture of \sysName}
\label{sec:arch}


In the off-line phase, we preprocess the data and instantiate our set-based model. Equi-depth binning is applied to each attribute and we use the LCM closed frequent pattern mining algorithm \cite{uno2004lcm} to generate \reva{(possibly overlapping)} itemsets. Different Reinforcement Learning models are trained as explained in Section \ref{subsec:rlsum}.
In the online phase, we allow users to generate summarization pipelines following one of the modes: {\em Manual} where the system displays a summary at each step, and the user inputs the next itemset, operation and corresponding attributes to be applied to the chosen itemset; {\em Partial Guidance} where at each step, the system displays a summary and the the user may provide only part of the input for the next step (e.g., specifying solely the target itemset, or the operator to apply);
{\em Full Guidance} where the system displays a $t$-size summarization pipeline. 
 Full and
partial guidance rely on executing a summarization pipeline. 
 
Pipeline execution starts by running the SWAP algorithm~\cite{swap} that finds the $k$ most uniform and diverse itemsets. 
Thus, for a summarization pipeline of length $1$, \sysName\ behaves exactly as the SWAP algorithm. 
The next steps executes one of \topsum\ or \rlsum. 
The algorithm picks one of the itemsets returned by SWAP or the previous operator, and chooses which operator to execute on that itemset, resulting in a new summary. \topsum\ is a simple greedy-based algorithm that at each step chooses to apply an operator which results in the summary with highest utility.
\rlsum\ auto-generates summarization sessions using Deep Reinforcement Learning (DRL). This solution allows us to reduce computation time at runtime. The models are pre-trained, and the inference time to pick the best expected action is insignificant. 
In our experimental study we compare the results of \topsum\ and \rlsum.




\subsection{\topsum}
\label{subsec:top1sum}
The \topsum\ algorithm applies local optimization to find the operation that produces the highest utility summary at each step of the summarization pipeline. Intuitively, at each step, \topsum\ examines every possible next step, i.e., every (itemset, $explore()$, attributes) combination (where the itemset is one of the itemsets the user is currently seeing), and executes the step that yields the summary with the highest utility. \revb{Formally, at every step, given a summary $I$, \topsum\ chooses the summary $I'$ s.t.:\\ $I' {=} argmax_{i \in I}utility(explore(i))$, where $explore()$ is the operator applied on the itemset $i {\in} I$ which results with the highest utility among all operators and input itemsets. }
%


\topsum\ has no theoretical guarantees for the \pbm\ Problem.
Nevertheless, as our experimental study shows, \topsum\ works well in practice, and it is able to generate high utility summarization pipelines.  
We note that the main drawback of \topsum\ is its running times, which are relatively slow even if the itemset vectors are precomputed. The utility computation of next-step summaries could be parallelized to speed up computation.



\subsection{\rlsum}
\label{subsec:rlsum}
Following recent approaches that were proven to be useful for EDA \cite{ATENA20,DBLP:journals/pvldb/SeleznovaOAS20,DBLP:conf/sigmod/PersonnazABFS21, DBLP:conf/cikm/PersonnazABFS21}, we present a Deep Reinforcement Learning solution to find a high-utility summarization pipeline. 

We model the \pbm\ Problem as a Markov Decision Process (MDP) comprising a triple ($S$, $E$, $R$) where:

\begin{compactitem}
    \item $S$ is a set of summarization states. Each state $s_i$ contains several itemsets referred to as $sets(s_i)$;
    \item $E$ is a set of actions, where each action is a specific exploration function $explore(x, a)$ (with $x$ an input itemset picked from $sets(s_t)$, and $a$ an optional attribute) and enables a transition between consecutive exploration states.
    \item $R(s_t,e_t,s_{t+1})$ are rewards for transitioning from state $s_t$ to $s_{t+1}$ by applying exploration action $e_t$.
\end{compactitem}

We define a \textit{summarization policy}~$\pi$ as a mapping function from an summarization state $s_t$ to an action $e_t$, where $\pi(s_t) = e_t$, and look for the policy maximizing expected reward such as:
\setlength{\abovedisplayskip}{0pt}
\setlength{\belowdisplayskip}{0pt}
\begin{equation}
\pi^{*}= argmax_{\pi}\mathop{{}\mathbb{E}}\Big[ \sum_{t=1}^{|\pi|} \gamma^{i} R(s_t, e_t, s_{t+1}) \Big]
\end{equation} 
where $\gamma$ is a discount factor in $[0,1]$, $|\pi|$ is the length of policy $\pi$.

We model the reward of the action $e_t$ on the state $s_t$ as the utility value of $sets(s_{t+1})$.
The \rlsum\ algorithm finds a policy $\pi^{*}$ that maximizes the expected cumulative reward.


There are many methods for solving MDPs, including value iteration and policy iteration. It has been proved theoretically and empirically in~\cite{pashenkova1996value}  that policy iteration is computationally more efficient and requires a smaller number of iterations to converge. 
Here we adapt model-free RL~\cite{kaelbling1996reinforcement,DBLP:journals/pvldb/SeleznovaOAS20,bar2020automatically,sutton2018reinforcement} with inputs $(\mathcal{S},E,R)$ as a policy iteration method which fits our proposed problem remarkably well in the absence of logs as training samples.


{\bf Deep Reinforcement Learning Algorithm.}
Model-free RL allows us to address the problem of finding a policy, i.e., a pipeline, that maximizes the discounted cumulative reward. 
 Actor-critic methods combine policy gradient methods with a learned value function. Each learning episode contains {\em action probabilities and values} that get periodically updated as the agent learns from the environment based on the reward function. The policy (the actor) adjusts action probabilities based on the current estimated advantage of taking that action; the value function (the critic) updates this advantage based on the rewards such as: 
$\mathit{Advantage}(s_i,e_i,s_{i+1}) \approx  R(s_i,e_i,s_{i+1})
 + \gamma V(s_{i+1}) - V(s_i)$ where $V(.)$ is the expected reward function.
Several workers run in parallel 
and update the actor and critic values. 
\revb{We train a worker as follows. We instantiate the environment interface for the worker with the utility weights defined for the training. For every operator execution step, an action is selected and executed, and the reward is computed. The value network learns a baseline state value to which the current reward estimate is compared to obtain the “advantage”. The policy network adjusts the log probabilities of the actions based on the advantage via the RL algorithm. We then train the policy with the newly computed advantage values and train the value function with the obtained reward. This process is completed in parallel by each worker. }


%% file: exp.tex
\section{Experimental Study}
\label{sec:exp}

This section presents experiments that evaluate the effectiveness and
efficacy of \sysName. We aim to address the following research questions. \textbf{Q1}:
How do \topsum\ and \rlsum\ compare to each other and to SWAP w.r.t. the utility of found summaries? \textbf{Q2}: What is the response time of our algorithms? 
\textbf{Q3}: Is guidance in generating summarization pipelines needed? 


\subsection{Implementation Details}
\label{sec:impl}


Our code is available at \cite{gitEDA4SUM}.
\common{\sysName\ is available at \cite{UI}}.
\revc{All algorithms are implemented in Python 3.7. The experiments were executed on a PC with i7-9850H 2.6GHz and 16GB Ram memory.}

We measure utility using the standard deviation measure.
In case some attributes are categorical, other deviation measures, such as entropy, could be used without affecting our solution. 
To measure diversity, we use the Manhattan distance metric as the vector distance measure.
Other vector distance metrics could be used with only minor modifications. 
To obtain comparable values, 
we used the scaling method that was presented in \cite{milo2019}.

\vspace{1mm}
\textit{Evolving weights}:
There is a trade-off between uniformity, diversity, and novelty of a summary. Finding a highly uniform and diverse summary may come at the cost of returning a novel one. In different parts of the summarization pipeline, the user's preferences may change. Suppose the user has seen in previous steps many itemsets. In this case, it is more important to return a uniform and diverse summary rather than a novel one. To capture this, we tested two evolving weights schemes: \emph{Increasing Novelty} and \emph{Decreasing Novelty}.
For these schemes, the novelty weight is a function of the total number of itemsets, the number of seen itemsets, and the length of the pipeline.  
This weighting scheme will be compared to others, such as fixed-value weights (e.g., balanced weights). \reva{To safe space, we do not report the results of increasing or decreasing uniformity (resp., diversity), as decreasing novelty is almost the same as increasing or decreasing uniformity (resp. diversity).}

\paragraph*{\rlsum\ implementation}
We use a Tensorflow-based implementation of A3C
~\cite{DBLP:conf/icml/MnihBMGLHSK16} as a policy learning method, a state-of-the-art DRL framework that has been shown to outperform other critic-based methods on a wide range of applications~\cite{DBLP:conf/icml/MnihBMGLHSK16}.
The appeal of A3C comes from its parallelized and asynchronous architecture: multiple actor-learners are dispatched to separate instantiations of the environment; they all interact with the environment and collect experience and asynchronously push their gradient updates to a central target network. 
\revc{The agents were trained on two servers with Intel Xeon processors, with 370GB and 126GB of RAM.
Training took $100$ hours for $4000$
episodes on SDSS and $3000$ on SPOTIFY \reva{and FOOD}, with $50$ steps per
episode. Each agent used 6 workers in parallel; the update interval
was set to 20 steps, and
we concatenated three successive states for the LSTM
layers.}



\paragraph*{Operators}
We support the following exploration operators:\\ 
\textbf{(1) \byfacet(i,a)} (drill-down): returns as many subsets of $i$ as there are combinations of values for the attribute $a$;
\textbf{(2) \bysuperset(i)} (roll-up): returns the $k$ smallest supersets of the input itemset $i$;
\textbf{(3) \bydistrib(i)}: returns $k$ itemsets whose attribute value distribution is similar to $i$;
\textbf{(4) \byneighbors(i,a)}: returns $2$ itemsets that are distinct from the input itemset $i$ and that have the previous (smaller) and next (larger) values for attribute $a$.

The \bydistrib\ and \byneighbors\ operators were introduced in~\cite{DBLP:conf/sigmod/PersonnazABFS21,DBLP:conf/cikm/PersonnazABFS21}
\reva{Given an itemset $i$, the \byneighbors\ operator only changes the value of one attribute in the description of $i$ to obtain the neighboring itemsets, and \bydistrib\ returns itemsets having similar distributions to $i$. Thus, if applied on a uniform itemset, the \bydistrib\ and \byneighbors\ operators would return itemsets that are (almost) as uniform as the input itemset. Also, by definition, both the \byfacet\ and the \bysuperset\ operators return itemsets having different descriptions and items than that of the input itemset. Thus, it is more likely to find a diverse summary after applying one of these operators than the \bydistrib\ and \byneighbors\ operators. Our experiments will examine those hypotheses. }

\subsection{Experimental Setup}
\label{subsec:setup}
\begin{table}
	\centering
	\captionsetup{justification=centering}	
	\small
		\caption{Examined Datasets.}
			\label{tab:data}
			\vspace{-12px}
	\begin{tabular} {|p{9mm}|p{10mm}|p{10mm}|p{16mm}|p{18mm}|}
		\hline
	\textbf{Dataset} & \textbf{|items|}& \textbf{|atts|}&\textbf{|itemsets|}&	\textbf{|ground truth itemsets|} 
	 \\
		\hline
	SDSS &$2.6$M&$7$&$348,857$&$169$\\
	  SPOTIFY &$232,725$&$11$&$2,204,806$&$27$ \\
	
				 \reva{FOOD} &\reva{11, 762}&\reva{11}&\reva{226, 381}&\reva{22}\\
				\hline
	\end{tabular}
	\vspace{-3mm}
\end{table}

\begin{figure}
    \centering
    \includegraphics[width=0.98\linewidth]{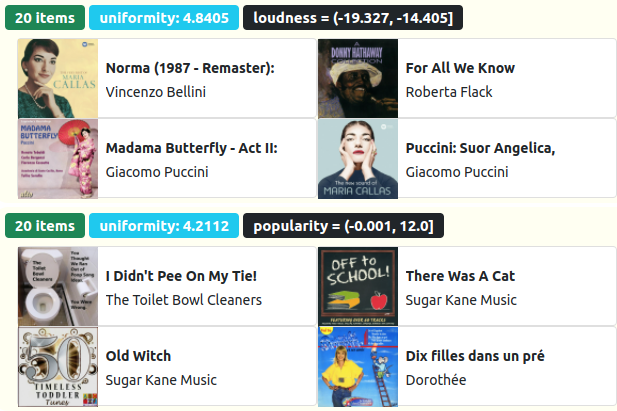}
    \vspace{-4mm}
    \caption{\revb{Example of song itemsets.}}
    \label{fig:example-spotify}
\end{figure}

To demonstrate the applicability of \sysName\ to multiple scenarios, we consider the following datasets. In all datasets, each column was binned into
$10$ equi-depth bins.
\textbf{SDSS.} 
SDSS \cite{SDSS}
is a massive sky survey dataset containing images
and metadata of astronomical objects. 
We selected $2.6$M galaxies with clean
photometry and spectral information. Each galaxy has $7$ attributes commonly used in Astronomy to describe information such as the magnitude in each color filter, and the size of an object.
\textbf{SPOTIFY Track DB.}
The SPOTIFY dataset is a publicly available Music database \cite{spotify}. It contains information about over $232$K music tracks. We have extracted the $11$ numerical attributes that describe the track valence, duration, danceability, etc. 
\revb{Figure~\ref{fig:example-spotify} shows a screenshot of \sysName\ with SPOTIFY songs.}
\reva{\textbf{FOOD Data.} The FOOD Data is a publicly available food nutrition facts dataset \cite{food}. It contains information about over $14$K food items. We have extracted $11$ numerical attributes that describe the item's amount of calories, fat, protein, etc. }

\paragraph*{Itemsets}
To build itemsets we use LCM, an implementation of the Apriori algorithm for frequent pattern mining~\cite{uno2004lcm}. 
Each frequent pattern is described with attributes which are common to all items of the pattern. Hence each pattern forms an itemset~$s$ where $s.desc$ is the pattern itself.
Here we used LCM with a support value of $10$ to generate 
$348,857$ itemsets whose size ranges from $10$ to $261,793$
galaxies (for SDSS), and with a support value of $20$ to generate $2,204,807$ itemsets whose size ranges from $20$ to $93,107$ music tracks (for SPOTIFY). \reva{For FOOD, we used a support value of $10$ to generate 
$226,381$ itemsets whose size ranges from $10$ to $4,953$.}

\paragraph*{Ground-truth uniform itemsets} For each dataset, we define a set of \reva{(non-overlapping) uniform itemsets to be discovered by a summary}. Those are referred to as "ground-truth" itemsets. For SDSS, they correspond to $169$ well-known galaxy types extracted from the Galaxy Zoo Classification \cite{Willett2013}, \reva{covering less than 12\% of the data}. 
For SPOTIFY, they correspond to the partition of all music records by the attribute genre. The number of genres
is $27$. \reva{For FOOD, the ground truth itemsets correspond to the partition of all food records by the attribute food group. The number of food groups is $22$}.
Importantly, we note that those itemsets are not necessarily the most uniform itemsets in each dataset.
Other ground-truths could be defined. We will see in Section~\ref{subsubsec:exe} how relevant the summaries we return are to our ground-truths.



Unless otherwise indicated, we set the number of steps to $50$ and the maximal size of a summary to $10$.  

\paragraph*{Variants}
\underline{One-shot summarization}. We implemented the common \textbf{SWAP} algorithm~\cite{swap}. 
The output of this algorithm is also the starting point of \topsum\ and \rlsum.
\underline{Multi-step summarization}. 
\textbf{Top1Sum} The greedy algorithm described in Section \ref{subsec:top1sum}.
\textbf{RLSum}. The RL-based algorithm described in Section \ref{subsec:rlsum}. \common{To compare \rlsum\ with existing EDA solutions, we included two additional RL-based baselines introduced in \cite{DBLP:conf/sigmod/PersonnazABFS21}: \textbf{FAMO}. A familiarity-only algorithm that mimics exiting EDA approaches, and \textbf{75FAM-25CUR} for 75\% familiarity and 25\% curiosity. This algorithm achieves the best results for the EDA task presented in \cite{DBLP:conf/sigmod/PersonnazABFS21}.}
\underline{Manual}. At each step, the user specifies the chosen itemset, operator and parameters, and the resulting summary is returned. 
\emph{This baseline serves to demonstrate the need for guidance in generating useful summaries}.

\reva{
For both \topsum\ and \rlsum\ we have experimented with different weighting schemes (fixed or evolving weights) for the parameters $\alpha,\beta$ and $\gamma$. We indicate high, balanced and low weight of a parameter with a suffix. For example, \rlsumbl\ with balanced weights on utility, diversity and novelty, and \rlsumln\ with a low novelty weight. 
\rlsumic\ (resp., \rlsumdc) denotes \rlsum\ with an increasing (resp., decreasing) novelty weight.} 



\begin{figure}[htpb]
	\small
	\begin{center}
		\begin{minipage}[b]{0.5\textwidth}
		\includegraphics[width=\linewidth]{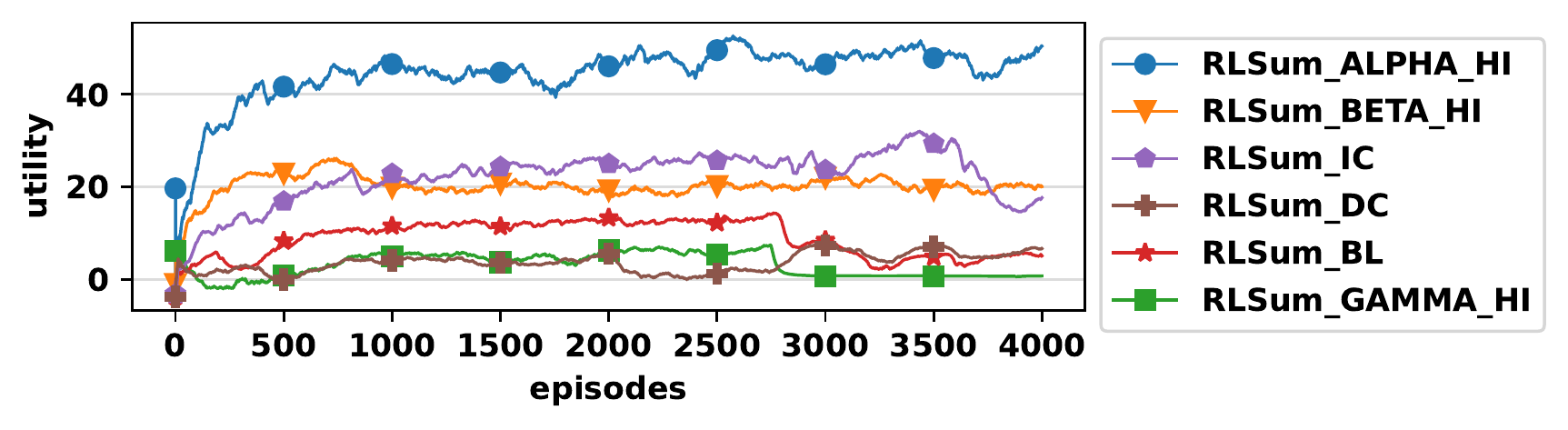}
		\vspace{-26px}
		\caption*{{SDSS}}  
	\end{minipage}%

		\end{center}
	\vspace{-14px}
	\caption{Cumulated utility during training.} \label{fig:utility_training}
	\vspace{-4mm}
\end{figure}

\subsection{Utility and Relevance to Ground-Truth (Q1)}
\label{subsec:exp_utility}
In what follows we set the uniformity threshold of SWAP to $2$.
We examined the impact of this threshold on utility over all datasets. We found that lower thresholds return itemsets that have poor uniformity while higher thresholds return too few itemsets. 

\paragraph*{Results Summary}
\common{We summarize our main finding as follows:\\
$\bullet$ As expected, the results clearly show that the \topsum\ variants produce
the highest utility values. \\
$\bullet$ In particular, \topsumhu\ achieves the highest cumulated utility in all datasets. This implies that to optimize utility, high weights for uniformity are required. \\
$\bullet$ However, in terms of quality, we see that different variants, particularly \rlsum\ variants, were performing better. This indicates that for different real-life scenarios, where the uniformity and diversity of the ground-truth itemsets varies, balancing uniformity, diversity, and novelty is required. \\
$\bullet$ This motivates the need for a tunable objective function, where users can set the balance among uniformity, diversity, and novelty.\\ 
$\bullet$ The results of FAMO and 75FAM-25CUR were inferior in terms of both utility and quality, showing that existing EDA solutions are ill-suited to our problem.}\\


\begin{figure*}[htpb]
	\small
	\begin{center}
		\begin{minipage}[b]{0.285\textwidth}
		\includegraphics[width=\linewidth]{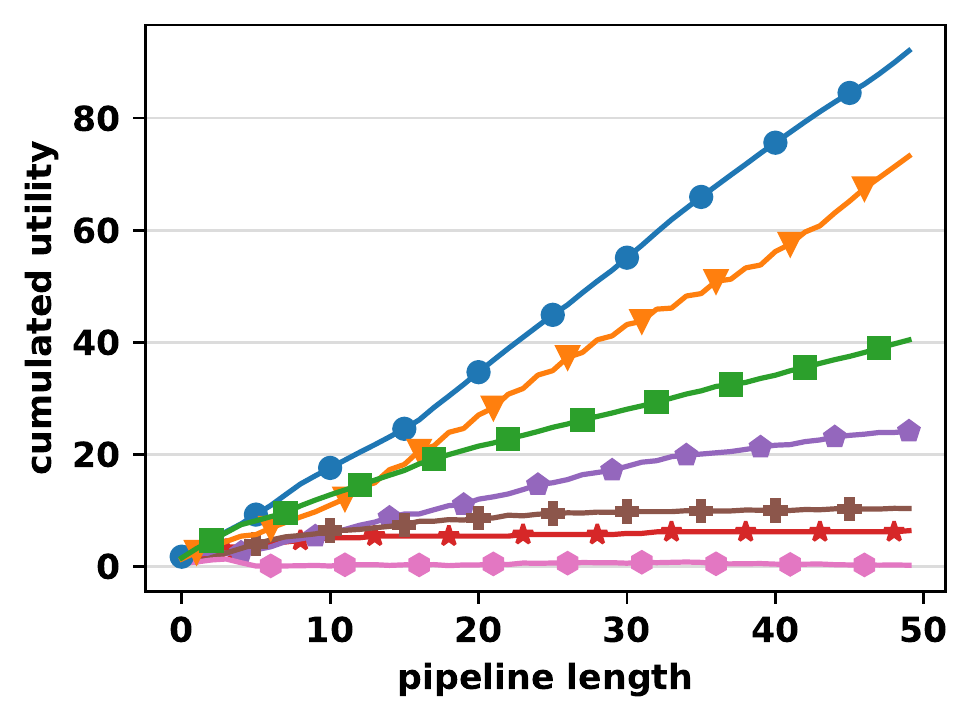}
		\vspace{-26px}
		\caption*{{SDSS}}  
	\end{minipage}%
	\begin{minipage}[b]{0.285\textwidth}
		\includegraphics[width=\linewidth]{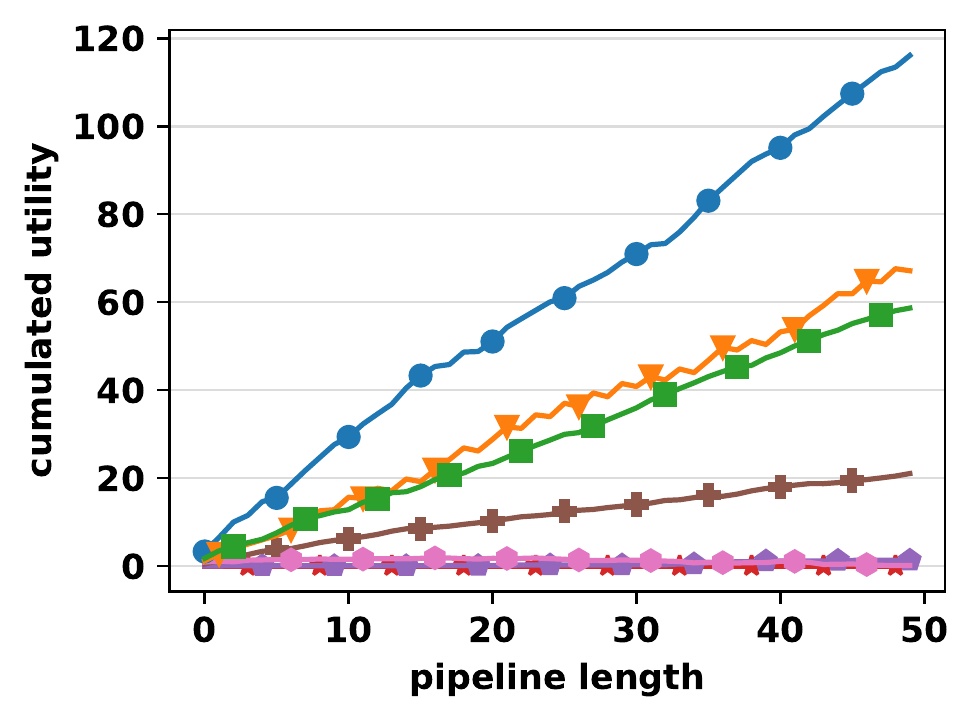}
		\vspace{-26px}
		\caption*{{SPOTIFY}}  
	\end{minipage}%
		\begin{minipage}[b]{0.43\textwidth}
		\includegraphics[width=\linewidth]{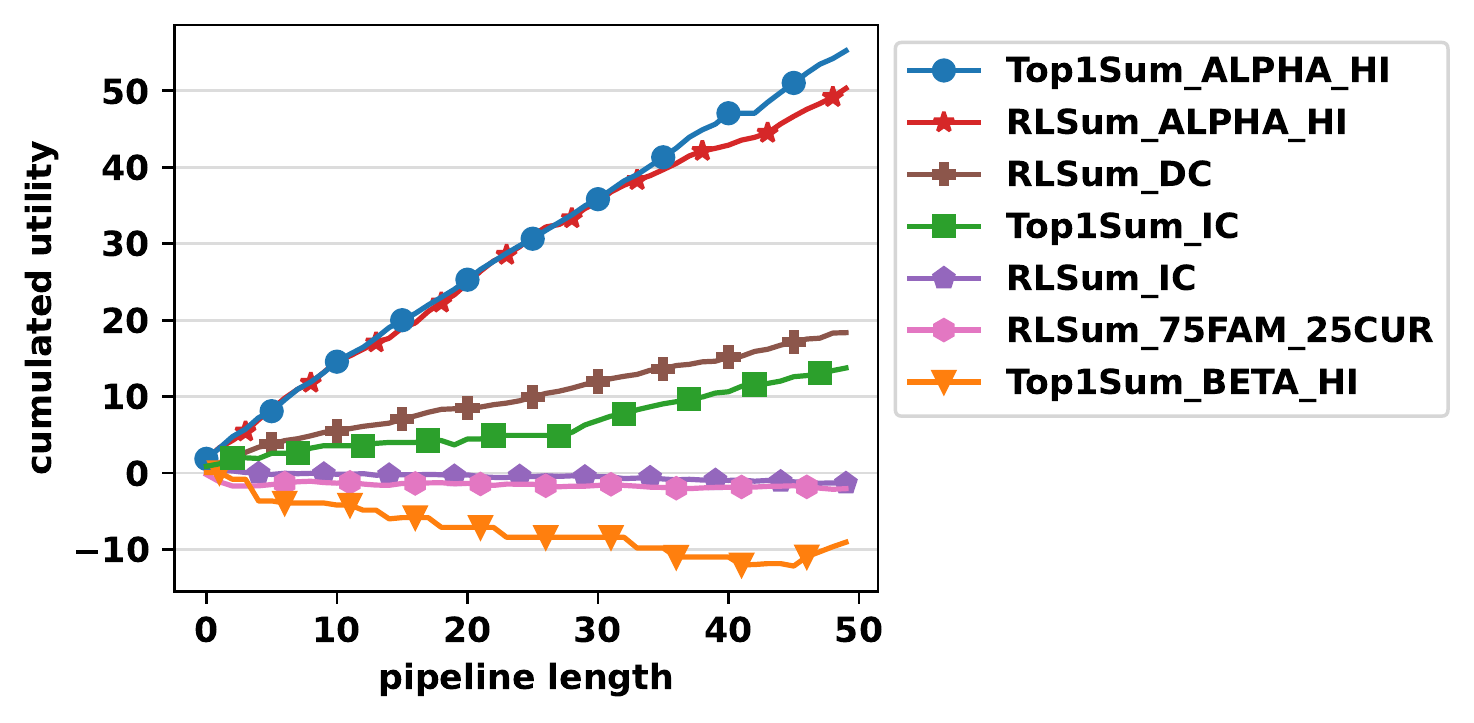}
		\vspace{-26px}
		\caption*{{FOOD}}  
	\end{minipage}%
		\end{center}
	\vspace{-14px}
	\caption{\common{Cumulated utility as a function of pipeline length.}} \label{fig:utility_progress}
	\vspace{-4mm}
\end{figure*}

\begin{figure*}[htpb]
	\small
	\begin{center}
		\begin{minipage}[b]{0.455\textwidth}
		\includegraphics[width=\linewidth]{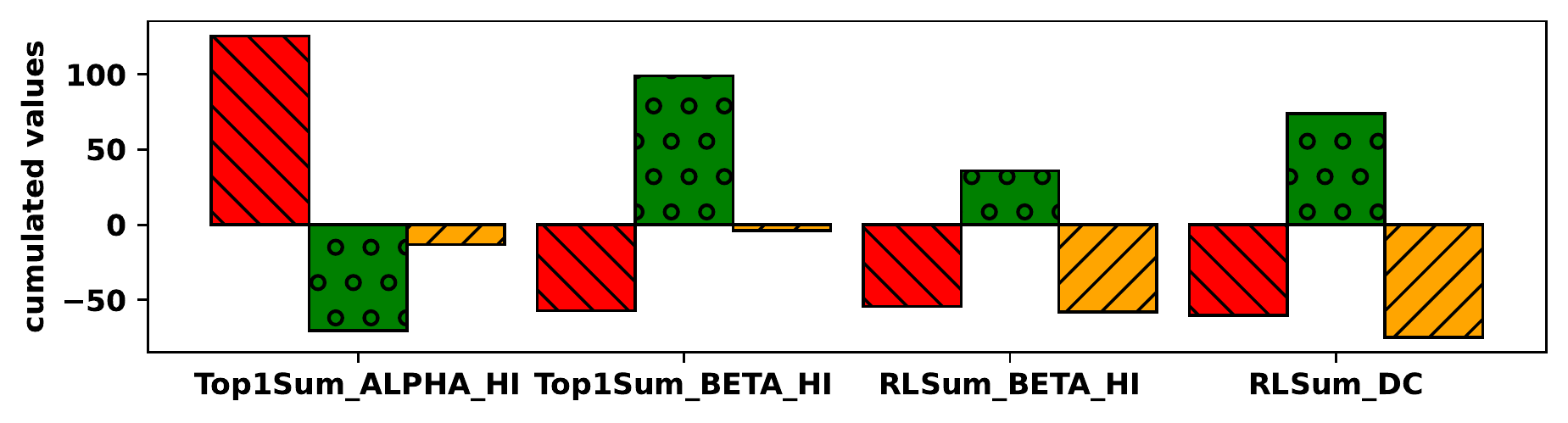}
		\vspace{-26px}
		\caption*{{SDSS}}  
	\end{minipage}%
	\begin{minipage}[b]{0.545\textwidth}
		\includegraphics[width=\linewidth]{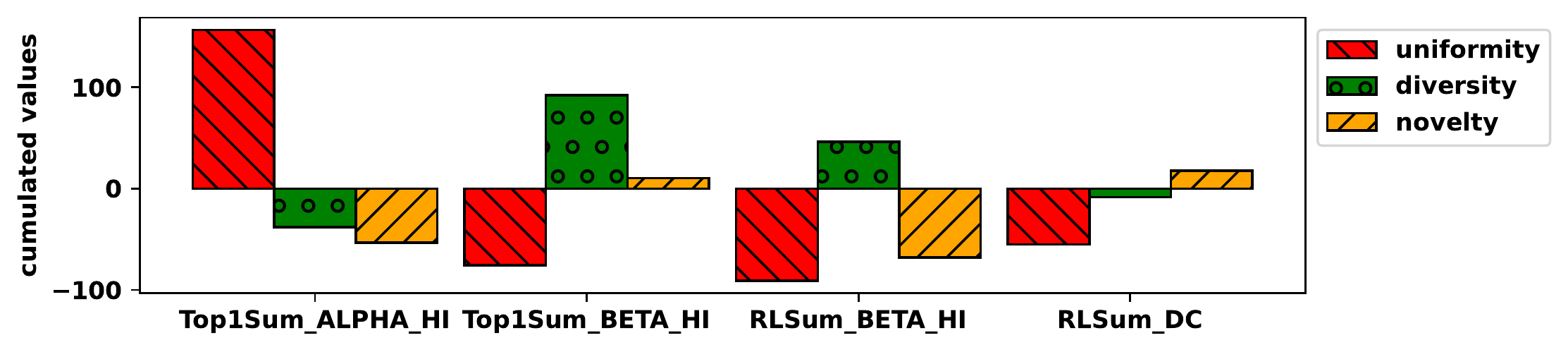}
		\vspace{-26px}
		\caption*{{SPOTIFY}}  
	\end{minipage}%
		\end{center}
	\vspace{-14px}
	\caption{\common{Cumulated uniformity, diversity and novelty during pipeline execution.}} \label{fig:uniformity_diversity_novelty}
	\vspace{-4mm}
\end{figure*}

\vspace{2mm}
\noindent
\textbf{Evolution of utility during training}.
Figure~\ref{fig:utility_training} reports the training of our policies for different \rlsum\ variants for SDSS. As can be seen, the agents are able to optimize and improve their utilities. 
Similar trends were observed on the other datasets as well. 

\subsubsection{Pipeline execution}. 
\label{subsubsec:exe}
We recorded data on a pipeline execution in the {\em Full Guidance} mode for each baseline algorithm. 

\vspace{2mm}
\noindent
\textbf{Utility during pipeline execution}. 
Figure~\ref{fig:utility_progress} shows the evolution of utility of the top-3 performing \rlsum\ and \topsum\ variants with pipeline length. Other variants in which their results were inferior were omitted from presentation. 
As expected, the results clearly show that \topsum\ variants produce the highest utility values. \common{The best performing variant in all datasets is \topsumhu. In Food, the smallest dataset, \rlsumhu\ was also managed to generate high utility pipelines, but note that this was not the case for the larger datasets. Not surprisingly, the results of the EDA solutions (FAMO and 75FAM-25CUR) are inferior, as these baselines are optimized for a different task. } 
We observe for all datasets that the difference between variants increases as the pipeline length increases.

We dive into this comparison and plot the values of utility dimensions: uniformity, diversity and novelty (Figure~\ref{fig:uniformity_diversity_novelty}) of the two best \topsum\ and \rlsum\ variants. These scores could take negative values due to our scaling procedure (see Section \ref{sec:impl}). The results on  FOOD showed similar trends. 
The first observation is that the weights assigned to these dimensions impact the performance of individual variants. This is quite apparent on \topsum\ variants where the value for each dimension reflects its weight (e.g., \topsumhu\ yields the highest cumulated uniformity). 
We observe that the \rlsum\ variants with a high novelty weight (e.g., \rlsumdc) achieve high diversity. Indeed, while novelty appears to be difficult to learn, trained agents compensate with other dimensions. This is appealing as it demonstrates the capability of \rlsum\ variants to adapt to the dataset. This will be studied further when we will examine the relevance of found summaries to a ground-truth.

\begin{figure*}[htpb]
	\small
	\begin{center}
		\begin{minipage}[b]{0.285\textwidth}
		\includegraphics[width=\linewidth]{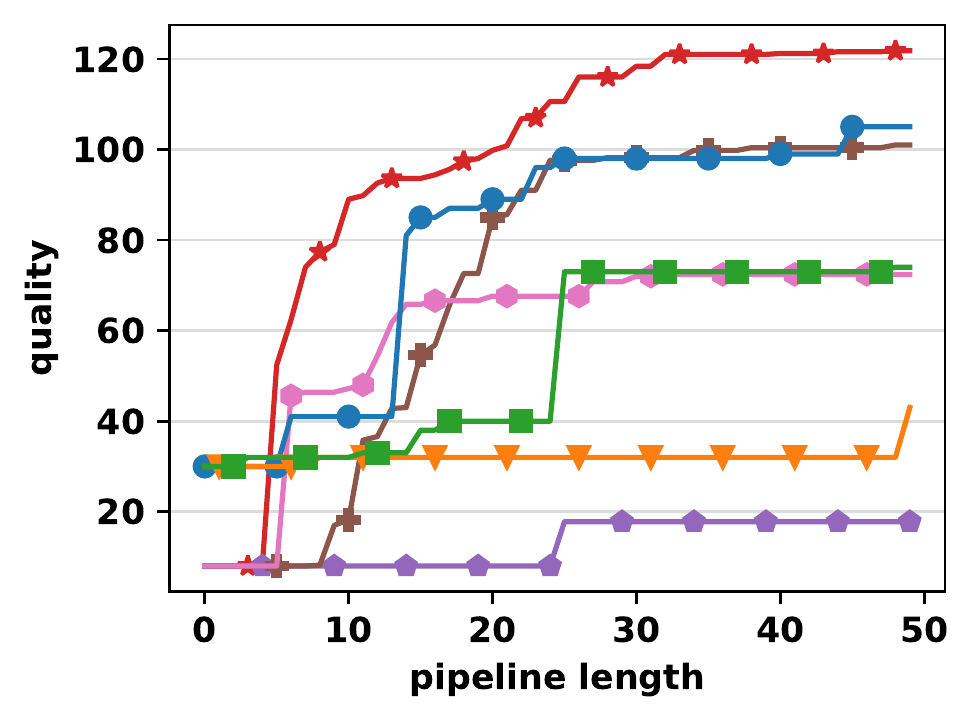}
		\vspace{-8mm}
		\caption*{{SDSS}}  
	\end{minipage}%
	\begin{minipage}[b]{0.285\textwidth}
		\includegraphics[width=\linewidth]{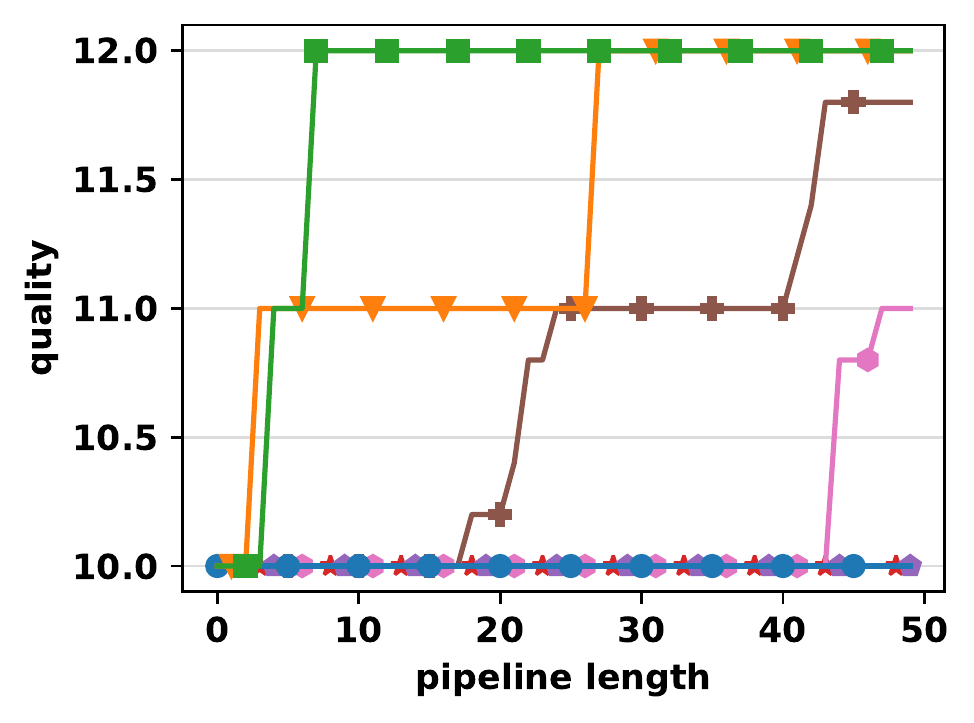}
		\vspace{-8mm}
		\caption*{{SPOTIFY}}  
	\end{minipage}%
		\begin{minipage}[b]{0.43\textwidth}
		\includegraphics[width=\linewidth]{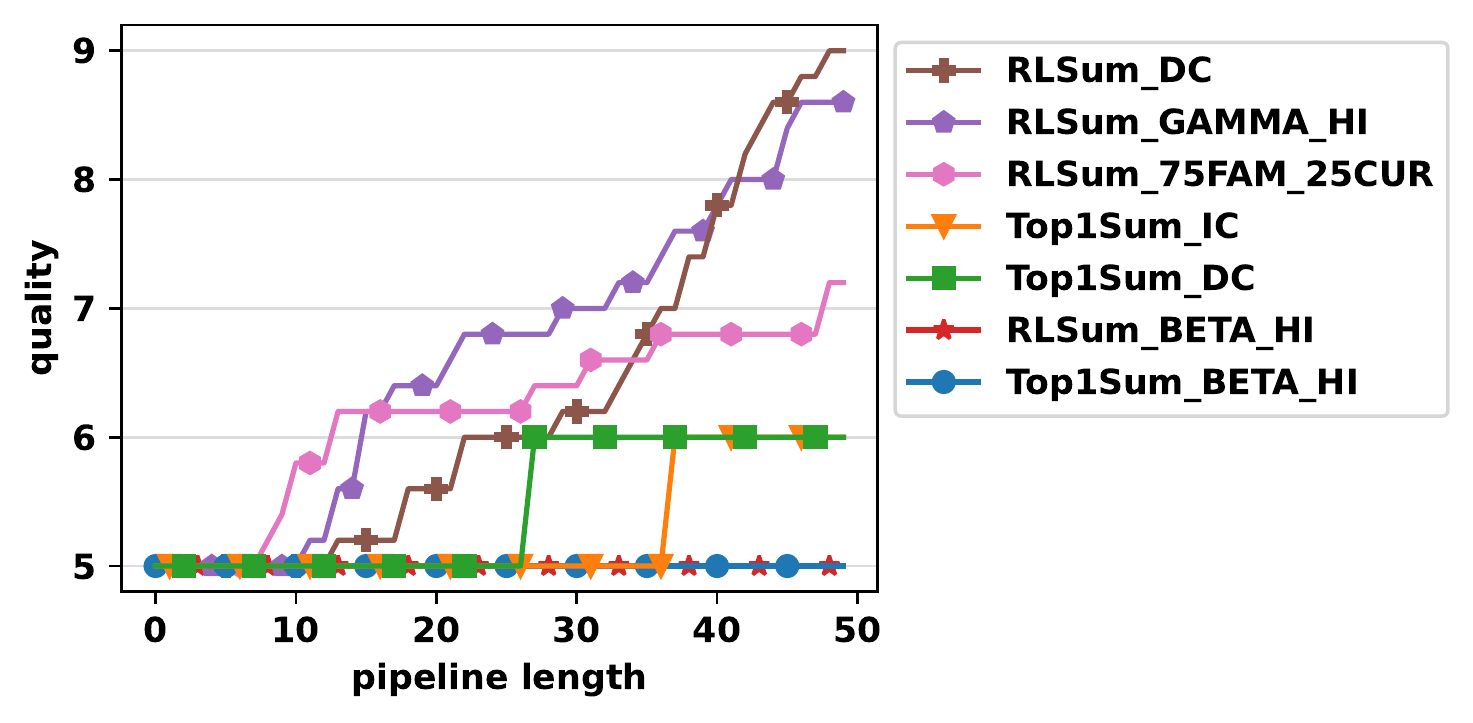}
		\vspace{-8mm}
		\caption*{{FOOD}}  
	\end{minipage}%
		\end{center}
	\vspace{-14px}
	\caption{\common{Cumulated relevance to a ground-truth as a function of pipeline length.}} \label{fig:relevance}
	\vspace{-4mm}
\end{figure*}

\vspace{2mm}
\noindent
\textbf{Discovering ground-truth itemsets}.
We now illustrate a use case that studies the relevance of obtained summaries, i.e., the number of discovered ground-truth itemsets. This clearly depends on the definition of ground-truth itemsets (e.g., in SPOTIFY these itemsets are not so uniform), and how similar they are to each other, which affects diversity.
Figure~\ref{fig:relevance} reports the cumulated relevance of itemsets found at each step during pipeline execution.
Here again, to ease the presentation, we plot only the results of the top-3 variants of \topsum\ and \rlsum\ that achieve the best results. 

The first observation is that the variants that achieve the highest results in terms of utility (e.g., \topsumhu) are not the ones with the highest relevance. \common{The explanation is that  ground truth itemsets are not necessarily the most uniform itemsets in the data. Furthermore, their diversity and uniformity levels vary across datasets. This warrants the need for a tunable objective where the user can set the weights of the utility dimensions according to her needs.} 



Our second observation is that despite \topsum\ outperforming \rlsum\ when measuring utility, in many cases, \rlsum\ outperforms \topsum\ when measuring relevance to a ground-truth.
We can see that \rlsumdc\ performed well on all datasets. 
The intuition is that, to reach high relevance summaries, we need to start with a high novelty weight and decrease it as the pipeline is executed. The observed relevance results and the utility dimensions obtained by \rlsumdc\ suggest that high relevance summaries depend on either a mix of diversity and novelty, or on a very high level of diversity. While uniformity eases interpretability, reaching representativity of large datasets requires to favor diversity and novelty, two dimensions that EDA is designed to optimize.
This experiment confirms the usefulness of EDA for summarizing large datasets. 


\begin{figure*}[htpb]
	\small
	\begin{center}
		\begin{minipage}[b]{0.455\textwidth}
		\includegraphics[width=\linewidth]{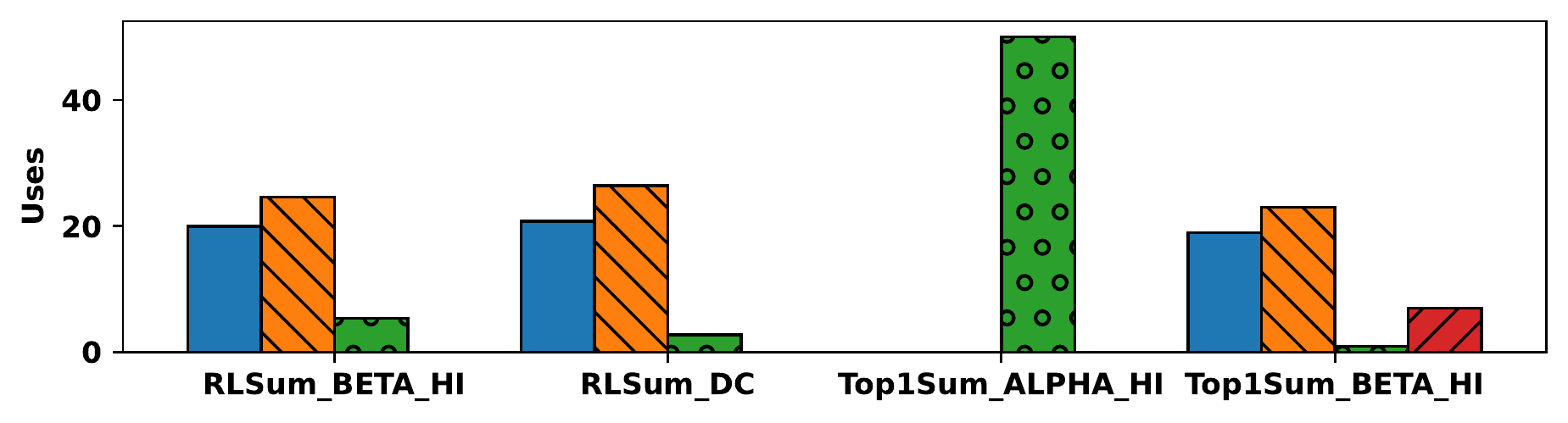}
		\vspace{-24px}
		\caption*{{SDSS}}  
	\end{minipage}%
	\begin{minipage}[b]{0.545\textwidth}
		\includegraphics[width=\linewidth]{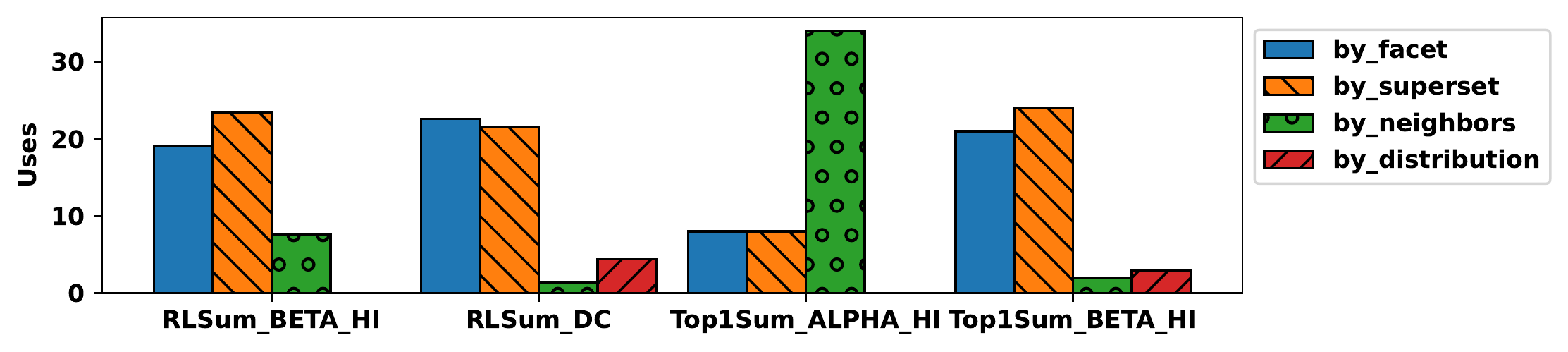}
		\vspace{-24px}
		\caption*{{SPOTIFY}}  
	\end{minipage}%
		\end{center}
	\vspace{-18px}
	\caption{\common{Operator usage in pipeline execution.}} \label{fig:operator_usage}
	\vspace{-2mm}
\end{figure*}

\vspace{1mm}
\noindent
\textbf{Impact of EDA operators}.
We examined the impact of using \sysName\ with 2OP (\byfacet\ and \bysuperset) vs. using all four operators. \common{We report that in all cases, the results of the 2OP versions of both \topsum\ and \rlsum\ were inferior to those achieved with all operators. This verifies the need in supporting expressive exploration operators that go beyond traditional drill-down and
roll-up.}
\reva{Figure~\ref{fig:operator_usage} reports the proportion of usage of each operator by each variant. Observe that \rlsumdc\ that attains high relevance on all datasets, mostly uses \byfacet\ and \bysuperset\ that favor diversity and encourage finding new itemsets.
Whereas \topsumhu\ that attains high utility, mostly uses \byneighbors\ that favor uniformity. This is confirmed by Figure \ref{fig:uniformity_diversity_novelty}, where we can see that \rlsumdc\ achieves high diversity and \topsumhu\ achieves high uniformity. }


%
%

\subsection{Scalability Evaluation (Q2)}
The running time of a single step is measured between the time an operation is picked and the time a summary is displayed. All pipelines are executed in {\em Full Guidance}, and we report the average of $5$ executions. 
Table~\ref{tab:scalability} reports results on SPOTIFY. \reva{Other datasets demonstrated similar trends}. \common{The running times of the EDA variants are the same as \rlsum, and thus omitted.}
We compare between only two variants \topsumhu\ and \rlsumdc\ (as the weights do not affect running times). As expected, the results clearly show that \rlsum\ outperforms \topsum\ by one order of magnitude and that the difference between the two increases with data size, \# attributes, and \# bins. 
Since \topsum\ checks every itemset against each possible next operator to determine the highest utility results, its execution time depends on the number of itemsets returned by each operator. Increasing the number of bins increases resulting itemsets (e.g., more facets), leading to higher execution times. Interestingly, the performance of \rlsum\ improves with a higher number of bins, as the number of mined itemsets reduces.
These results confirm that \rlsum\ is the method of choice for performing interactive summarization. Obviously, \topsum\ returns the highest utility summaries and may still be preferred given that \rlsum\ comes at the cost of a long training time. 

\begin{table}
	\centering
	\captionsetup{justification=centering}	
	\small
		\caption{Average pipeline execution times (in seconds).}
			\label{tab:scalability}
			\vspace{-12px}
	\begin{tabular} {|p{10mm}|p{4mm}|p{5mm}|p{5mm}|p{4mm}|p{4mm}|p{4mm}|p{4mm}|p{4mm}|p{4mm}|}
		\hline
	\multirow{2}{*}{\textbf{Variant}} &\multicolumn{3}{c|}{\textbf{Data size}}&\multicolumn{3}{c|}{\textbf{\# of attributes}}&\multicolumn{3}{c|}{\textbf{\# of bins}}
	 \\
	 \cline{2-10}
	 &\textbf{23K}&\textbf{115K}&\textbf{233K}&\textbf{3}&\textbf{7}&\textbf{11}&\textbf{5}&\textbf{10}&\textbf{20}\\
		\hline
			\textbf{\topsum} &$4.3$&$17.8$&$21.1$&$0.6$&$4.1$&$21.1$&$19.7$&$21.1$&$32.8$\\
	  \textbf{\rlsum} &$0.4$&$0.7$&$1.1$&$0.4$&$0.5$&$1.1$&$1.4$&$1.1$&$0.8$ \\
	\hline
	\end{tabular}
	\vspace{-4mm}
\end{table}

\begin{table}
	\centering
	\captionsetup{justification=centering}	
	\small
		\caption{User study.}
			\label{tab:user_study}
			\vspace{-12px}
	\begin{tabular} {|p{20mm}|p{11mm}|p{8mm}|p{7mm}|p{10mm}|p{8mm}|}
		\hline
	\textbf{Mode} & \textbf{|itemsets|} & \textbf{utility} & \textbf{uni.} & \textbf{diversity} & \textbf{novelty}
	 \\
		\hline
Manual&$67$&$7.34$&$16.97$&$-0.5$&$3.26$\\
Partial Guidance&$142$&$2.35$&$-15.45$&$39.9$&$-24.77$\\
Full Guidance&$101$&$10.6$&$-59.51$&$68.07$&$-71.58$\\
				\hline
	\end{tabular}
	\vspace{-4mm}
\end{table}

\subsection{Summarization Guidance (Q3)}
To examine the benefit of guidance during summarization, we ran a user study and compared among all summarization modes. 

We asked two astronomers from the Max Planck Institute of Astrophysics and who are highly familiar with SDSS, to use \sysName. The request was to find as many well-known galaxies as possible from the $169$ galaxy types in Galaxy Zoo. The pipeline length was fixed to $50$.
The first astronomer was asked to use \sysName\ with the \emph{Manual} mode, and the second used the \emph{Partial Guidance} mode (with \rlsumdc). We recorded their pipelines and compared the results with a pipeline generated by {\em Full Guidance}. Table \ref{tab:user_study} report for each pipeline, the number of ground-truth itemsets discovered,
as well as their cumulated utility and its three dimensions.

We observe that \sysName\ with partial guidance outperforms the other two modes. \reva{Additionally, our astronomers clearly favored the sequence of connected itemsets over a set of itemsets.} 

\begin{figure}
    \centering
    \includegraphics[width=0.9\linewidth]{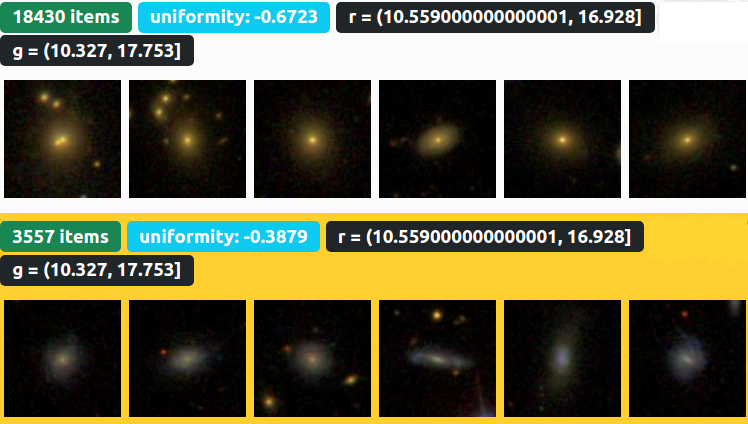}
    \vspace{-3mm}
    \caption{\revc{A relevant summary found with partial-guidance.}}
    \label{fig:userstudy}
    \vspace{-4mm}
\end{figure}

Our experts found the use of a system that encapsulates SQL and provides a visual interface very convenient. Hence, their preferences go to the manual mode as it lets them keep control over the summarization process. However, as shown in 
Table \ref{tab:user_study}, partially-guided pipelines yield the highest relevance by far.  
Hence, an expert with partial guidance manages to find almost all of the ground-truth itemsets in only fifty steps. Figure~\ref{fig:userstudy} illustrates one of the best summaries found with partial guidance. 
A deeper dive into utility shows that in the manual mode, experts find highly uniform itemsets with some reasonable novelty as they judiciously choose operators that do not rediscover seen itemsets. However, they are not able to find very diverse itemsets. As a result, relevance is the lowest. Interestingly, full-guided pipelines (with no expert intervention) yield the highest diversity, while partial-guidance allows them to \reva{connect summaries} and control the level of novelty and uniformity while improving on the diversity of the manual mode. This suggests that some automation with expert intervention is useful. 

%% file: conc.tex
\section{Conclusion}
\label{sec:conc}

This work examined the applicability of EDA to data summarization. Intuitively, \emph{a useful summary contains $k$ individually uniform sets that are collectively diverse} to be representative. This bears similarity to the well-known diversity problem where the goal is to find a set of relevant and diverse items. This observation allows us to formulate a multi-step summarization problem that seeks to build a pipeline that returns the most useful summaries. We propose two algorithms that adapt existing solutions. We run extensive experiments that validate the use of DRL for data summarization. Future work would examine if tuning the hyper parameters of the DRL algorithm may improve results. 


\begin{acks}
This work is funded by the European Union’s Horizon 2020 research and innovation program (project name: INODE) under grant agreement No 863410. 
\end{acks}


\newpage